%% file: main.tex
\documentclass{preprint}

\title[Neural Differentiable ICBFs]{Neural Integral Control Barrier Functions for Agile Collision Avoidance}
\usepackage{times}
\input{packages}

\input{macros}

\author{%
  \Name{Dasari Raj Vamsi} \Email{imt\_2022035@iiitm.ac.in}\\
 \addr Indian Institute of Information Technology and Management, Gwalior
  \AND
  \Name{Vrushabh Zinage} \Email{vrushabh.zinage@caltech.edu}\\
 \addr California Institute of Technology
 \AND
 \Name{Efstathios Bakolas} \Email{bakolas@austin.utexas.edu}\\
 \addr The University of Texas at Austin%
 \AND
 \Name{Rohan Chandra} \Email{rohanchandra@virginia.edu}\\
 \addr University of Virginia%
}

\begin{document}

\maketitle
\begin{abstract}
Control barrier functions (CBFs) are a standard way to keep a planner inside a safe set, but they assume control affine dynamics, a barrier function given in closed form, and a known nominal controller. Integral control barrier functions (ICBFs) remove the first assumption and absorb input constraints, yet they still assume a given nominal controller and a barrier function. In this work, we present an approach to learn all three from data, using neural ordinary differential equations (ODEs) for the dynamics, imitation for the nominal controller, and supervised regression for the barrier function. The learned dynamics, nominal controller, and the barrier function are used to build the ICBF filter constraint analytically. Furthermore, the barrier is trained on viability rather than geometry, meaning that a state counts as safe only if some emergency maneuver can still avoid the constraint under the actuator limits, a correction that geometric labels miss entirely and that grows with speed. Actuator limits then enter the online program as hard bounds while the learned barrier rows may be relaxed, which keeps the program solvable at every state and the actuator limits exact, and we bound how far the input it returns can drift from the one the true models would give. On agile racing with a 6-D dynamic bicycle, the filter removes all 12 unsafe episodes from 200 held out layouts at an unchanged success rate and above $100$\,Hz on one CPU core, holds the input constraint on every episode of a benchmark where a jointly trained baseline holds it on $53.38\%$, and keeps that advantage under large errors in the plant friction.
\end{abstract}

\section{Introduction}

Autonomous vehicles \citep{hussain2018autonomous_cars}, factory robots, and multi robot teams \citep{beltrame2018engineering_multi_robot_systems_1} must respect hard state and actuator constraints while a performance objective drives them toward the limits of the hardware \citep{wabersich2023data_safety_systems}. The prevailing architecture for this is the safety filter: a nominal planner proposes an input, and a small optimization problem projects that input onto the set of inputs that keep a barrier function nonnegative. With CBFs this projection is a quadratic program (QP) that can be solved at control rates \citep{ames2019control_ames_cbf_start}.

The CBF-QP is normally stated under three assumptions. First, the dynamics must be control affine, since otherwise the filter becomes a sequence of nonlinear programs that are expensive to solve and difficult to analyze \citep{son2019safety_cbf_nlp_1,seo2022safety_cbf_nlp_2,agrawal2017discrete_cbf_nlp_3}. For example, a vehicle with saturating tire forces, a steering slew limit and motor lag is not control affine. Second, a valid barrier must be known in closed form. For an obstacle field it is usually taken to be a signed distance, but the set of states from which a collision can still be avoided is fixed by the dynamics and the actuator limits rather than by the geometry of the obstacles. Learned barriers such as \citet{srinivasan2020synthesis_svm_cbf} relax this requirement but remain restricted to control affine systems without input constraints. Third, a stabilizing feedback law $k(\bx)$ for the full nonlinear model is assumed, which is meant to be similar to the original problem. Input constraints compound all three, since the QP need not stay feasible once $\bu$ is confined to a compact set. Feasibility \citep{zeng2021enhancing_recursive_1,xiao2023feasibility_recursive_2,xiao2023learning_recursive_3,liu2023feasibility_recursive_4} and input constraints \citep{agrawal2021safe_input_constrained_1,breeden2021high_input_constraints_2,breeden2023robust_input_constrained_3,black2023consolidated_input_constrained_4} have both been treated, but under the above three assumptions.

ICBFs \citep{ames2020integral_cbf} resolve the first difficulty together with the input constraint. The input is lifted into the state, driven by an integral law $\dot{\bu}=\phi(\bx,\bu)+\bv$, and the QP solves for the auxiliary input $\bv$. The construction applies to general nonlinear systems and encodes state and input constraints in one scalar function. Current ICBF theory, however, assumes that $\phi$ (assembled from a nominal controller), the dynamics, and the barrier are defined apriori. The open question is: can ICBF-based safety be ensured for systems with \textit{unknown} dynamics, \textit{unknown} ICBF formulation, and an \textit{unknown} nominal controller? 

This paper proposes a modular approach that learns all three from data while keeping the ICBF construction analytic. We learn the state transition dynamics $F_\theta$ with a Neural ODE \citep{chen2018neural_ode}, a nominal controller $\pi_\theta$ by imitation of an unconstrained planner, and a state barrier $h_\theta(\bx)$ by supervised regression. Everything the filter needs is then differentiated out of those three networks, so the online cost is one small quadratic program per control step (Section~\ref{sec:main_result}), and no part of the safety requirement is learned as a penalty term. What decides whether such a filter is usable is not the learning itself but two choices inside the construction, which we study in detail.

At high speeds, a state outside an obstacle can also be unsafe, because braking distance and steering authority, not the shape of the obstacle, decide whether a collision is still avoidable. We therefore call a state safe only if at least one emergency maneuver, simulated through $F_\theta$ with the actuator limits in the loop, keeps the vehicle safe. Furthermore, actuator limits are physics and cannot be violated, and so, actuator range and rate limits become bounds on the decision variable.

\textbf{Main Contributions.}
\begin{enumerate}
\item A complete learning based realization of the ICBF safety filter for systems whose dynamics, nominal controller and barrier are all unknown, with every ICBF quantity obtained analytically from the three learned models rather than assumed (Section~\ref{sec:main_result}).
\item The online program is feasible and input invariant (Proposition~\ref{prop:feasible}). An error bound (Theorem~\ref{thm:app_error}) further shows that the input computed from the learned models approaches the one the same construction would produce from the true dynamics, controller, and barrier function, and identifies how much authority the input retains over the barrier as the quantity that governs the remaining gap.

\end{enumerate}

We evaluate on a kinematic vehicle and a quadrotor carrying a suspended pendulum, against SABLAS \citep{qin2022sablas} on its own double integrator benchmark, and on agile racing with a 6-D dynamic single track model with F1TENTH parameters \citep{okelly2020f1tenth}. An analytical HOCBF-QP \citep{xiao2022hocbf} written for a kinematic model that the vehicle does not obey is unsafe on $29$ to $35\%$ of episodes, because the barrier it enforces is certified for dynamics other than the ones in force. Filtering an MPPI planner \citep{williams2016mppi} with the learned ICBF at $4.0$\,m/s instead leaves 0 of 200 held out layouts unsafe against 12 for the planner alone ($p=0.0005$, paired exact McNemar test), at an unchanged success rate and with every parameter frozen before those layouts were generated.

\section{Related work\label{sec:related_work}}
\textbf{Input constraints and ICBFs.} Once $\bu$ is confined to a compact set, a CBF-QP can lose feasibility. Input constrained CBFs restrict the safe set to the part that bounded inputs can hold \citep{agrawal2021safe_input_constrained_1}, with extensions to high relative degree and robustness \citep{breeden2021high_input_constraints_2,breeden2023robust_input_constrained_3}, online adaptation \citep{black2023consolidated_input_constrained_4} and learned feasibility constraints \citep{xiao2023learning_recursive_3}, but all keep $\bu$ as the decision variable of a control affine model with a barrier in closed form. ICBFs \citep{ames2020integral_cbf} instead lift the input into the state through the integral law $\dot{\bu}=\phi(\bx,\bu)+\bv$, so that state and input constraints become constraints on one augmented system, the dynamics need not be control affine, and the program is solved for $\bv$. Later work made the construction robust to disturbances and extended it to higher relative degree through a disturbance observer \citep{zinage2023disturbance}, and handled input delay through predictor feedback with a closed form feasibility condition for joint state and input constraints \citep{kiss2026icbf_delay}. Throughout that line of work the dynamics, the barrier and the nominal controller inside $\phi$ are given in closed form, whereas we keep the ICBF construction unchanged and supply all three from data.

\textbf{Neural ICBFs.} Learning has entered ICBF based control only recently. Decentralized neural ICBFs for multi agent navigation \citep{zinage2024decentralized} learn $h_\theta(\bx,\bu)$ jointly with a policy by penalizing violations of the ICBF condition in the training loss, recover safety with an embedded MPC based ICBF, and assume that the agent dynamics and a nominal controller defining $\phi$ (an LQR or PID law) are known a priori. Neural CBFs for input limited systems outside the integral formulation, trained against saturating counterexamples \citep{liu2023safe} or as the value function of a nominal policy \citep{so2024train}, likewise assume a known control affine model. Our work differs from these in assuming none of the dynamics, the barrier or the nominal controller, in training the barrier on viability through the learned dynamics rather than on geometry or on a penalized barrier condition, and in enforcing actuator limits as hard bounds of the online program while only the learned barrier rows may be relaxed.

\textbf{Learned barriers, and what they are trained on.} Model based synthesis \citep{ames2019control_ames_cbf_start,zinage2023disturbance} requires the constraint set in closed form. Data driven alternatives replace it with a network trained from sampled states \citep{dawson2022safe_neural_cbf_4,taylor2020learning_linear_cbf_3,zinage2023neural_koopman_cbf,qin2021learning_neural_cbf_1,ahmadi2019safe_neural_cbf_2,yu2023learning_neural_cbf_3}, from expert trajectories \citep{robey2020learning_other_neural_cbf_1,boffi2021learning_other_neural_cbf_2}, or jointly with a policy \citep{qin2021learning_other_neural_cbf_3,meng2021reactive_other_neural_cbf_4,wang2021learning_other_neural_cbf_5}. Across this literature the labels are geometric, so the learned set is a smoothed obstacle indicator rather than a set from which the constraint can still be met, and that distinction is what viability labeling addresses. Episodic CBF learning \citep{cheng2019end_linear_cbf_1,wang2018safe_linear_cbf_2,taylor2020control_linear_cbf_4}, networks trained to emulate a CBF controller \citep{yaghoubi2020training_dnn_cbf} and BarrierNet \citep{xiao2021barriernet} all start from a valid CBF, and the latter two from known dynamics.

\textbf{Where safety is enforced.} SABLAS \citep{qin2022sablas} jointly learns black box dynamics and a control affine safety filter, with safety and input compliance expressed as penalty terms in the training loss, where the optimizer trades them against the task objective so that neither is met exactly. We keep the three learning problems separate and move safety into the online program, which makes each component independently verifiable, makes the error propagation of Theorem~\ref{thm:app_error} tractable, and removes the trade off between tracking and constraint satisfaction, a choice whose cost as well as its benefit is quantified in Section~\ref{subsec:discussion}. Imitation learning \citep{Kostrikov2019ImitationLearningViaOffPolicyDistributionMatching,Dadashi2020PrimalWassersteinImitationLearning,zheng2022imitation,bco,ifovideo} supplies the nominal controller but cannot encode state or input constraints, which is why the filter is needed at all.

\section{Preliminaries and Problem Statement\label{sec:problem_statement}}
Consider the general nonlinear controlled system
\begin{align}
    \dot{\bx}=F(\bx,\bu),\quad\quad \bx(0)=\bx^0
    \label{eqn:nonlinear_system}
\end{align}
where $\bx\in\mathcal{X}\subset\mathbb{R}^n$, $\bu\in\mathcal{U}\subset\mathbb{R}^m$, $F:\mathcal{X}\times\mathcal{U}\rightarrow\mathbb{R}^n$ is continuously differentiable, and $\mathcal{X}$ and $\mathcal{U}$ are compact. Let $b:\mathcal{X}\rightarrow\mathbb{R}$ satisfy $b(\bx)>0$ on $\text{int}(\mathcal{S}_\bx)$, $b(\bx)=0$ on $\delta\mathcal{S}_\bx$ and $b(\bx)<0$ on $\mathcal{X}\setminus\mathcal{S}_\bx$, where $\mathcal{S}_\bx\subset\mathcal{X}$ is the safe set.
\begin{definition}
     A function $b$ as above is a Control Barrier Function (CBF) if there exists a $\mathcal{K}_\infty$ function $\gamma$ such that
        $\underset{\bu\in\mathcal{U}}{\inf}\;\frac{\partial b}{\partial \bx}(\bx)\,F(\bx,\bu)\geq -\gamma(b(\bx))$
        for every $\bx\in\mathcal{X}$.
\end{definition}
If \eqref{eqn:nonlinear_system} is control affine, $F=f(\bx)+g(\bx)\bu$, safe inputs follow from the CBF-QP $\operatorname{argmin}_{\bu}\|\bu-k(\bx)\|^2$ subject to $L_fb+L_gb\,\bu\geq-\gamma(b)$, with $k(\bx)$ a stabilizing feedback controller. That program assumes control affine dynamics and a known $k$, and need not remain feasible under input constraints, both of which restrictions the integral formulation below removes.

\subsection{Integral Control Barrier Functions (ICBFs)\label{subsec:icbf}}
Consider the integral feedback law $\dot{\bu}=\phi(\bx, \bu)$ with $\bu(0):=\bu^0$, where $\phi: \mathcal{X} \times \mathcal{U} \rightarrow \mathbb{R}^m$ is continuously differentiable. Combined with \eqref{eqn:nonlinear_system} this gives
\begin{align}
[\dot{\bx}^\top ,\;\dot{\bu}^\top]
=[F(\bx, \bu)^\top,\;\phi(\bx, \bu)^\top] ,\; \bx(0)=\bx_0,\;\bu(0)=\bu_0.
\label{eqn:integrated_augmeted_system}
\end{align}
Let $\bz:=[\bx^{\top}, \bu^{\top}]^{\top}$ for the augmented state and $\mathcal{S}=\{(\bx,\bu):\;h(\bx,\bu)\geq 0\}$ for a safety set covering both state and input constraints. Forward invariance of $\mathcal{S}$ requires $\dot{h}(\bz)+\gamma(h(\bz)) \geq 0$ along every trajectory, which is enforced through an auxiliary input $\bv$ in $\dot{\bu}=\phi(\bx,\bu)+\bv$ \citep{ames2020integral_cbf}.

\begin{definition}
\label{def:icbf}
 Let $\gamma$ be a class $\mathcal{K}$ function. A continuously differentiable $h:\mathcal{X}\times\mathcal{U}\rightarrow\mathbb{R}$ with $\mathcal{S}=\{(\bx,\bu):\;h(\bx,\bu)\geq 0\}$ is an Integral Control Barrier Function (ICBF) if, for every $(\bx,\bu)\in\mathcal{S}$, $q(\bx, \bu) \leq 0$ whenever $p(\bx, \bu)=0$, where
$p(\bx, \bu)  :=\left(\frac{\partial h(\bx,\bu)}{\partial \bu}\right)^{\top}$ and
$q(\bx, \bu):=-\big(\frac{\partial h(\bx,\bu)}{\partial \bx} F(\bx, \bu)+\frac{\partial h(\bx,\bu)}{\partial \bu} \phi(\bx, \bu)+\gamma(h(\bx, \bu))\big)$.
\end{definition}
\begin{theorem}[\citealp{ames2020integral_cbf}]\label{thm:icbf}
Let $\phi(\bx,\bu)$ be an integral feedback controller and let $\mathcal{S}$ be defined by an ICBF $h(\bx,\bu)$ in the sense of Definition~\ref{def:icbf}. Modify the integral controller to $\dot{\bu}=\phi(\bx, \bu)+\bv^\star(\bx, \bu)$, where
\begin{align}
\bv^\star(\bx, \bu) = \underset{\bv \in \mathbb{R}^m}{\operatorname{argmin}}\|\bv\|^2\quad
\text {s.t. }\;\; p(\bx, \bu)^{\top} \bv \geq q(\bx, \bu).
\label{eqn:qp_icbf}
\end{align}
If the resulting closed loop admits a solution on the interval considered, then $\mathcal{S}$ is forward invariant along it for every initial condition $(\bx^0,\bu^0)\in\mathcal{S}$.
\end{theorem}
Following \citet{ames2020integral_cbf}, we do not learn a barrier over the joint state and input space. We learn a state barrier $h_\theta(\bx)$, form the extended barrier $h_{e,\theta}(\bx,\bu)=\dot h_\theta(\bx,\bu)+\gamma_x(h_\theta(\bx))$, and treat the actuator limits separately. The set $\mathcal{S}_e=\{(\bx,\bu):h_{e,\theta}(\bx,\bu)\geq0\}$ is therefore not a Cartesian product $\mathcal{S}_x\times\mathcal{U}$: it couples $\bx$ and $\bu$ through the dynamics and depends on how a given input moves the state. We now state the overall objective of this paper:

\begin{tcolorbox}[
  colback=gray!30,
  colframe=black,
  boxrule=0.6pt,
  arc=0pt,
  left=6pt, right=6pt, top=6pt, bottom=6pt
]
\begin{problem}
     Given that the dynamics $F$ of \eqref{eqn:nonlinear_system}, a valid ICBF, and the nominal controller from which the integral law $\phi$ is assembled are all unknown a priori, design an integral feedback law that enforces safety for \eqref{eqn:nonlinear_system} together with satisfaction of the input constraint $\bu\in\mathcal{U}$.
\end{problem}
\end{tcolorbox}

\section{Proposed approach\label{sec:main_result}}
We learn, in order, $(i)$ the dynamics $F(\bx,\bu)$ using neural ODE learning techniques, $(ii)$ a nominal controller in place of $k(\bx)$ using imitation learning, and $(iii)$ a state barrier $h_\theta(\bx)$, and combine them into the safe controller \eqref{eqn:u_safe_neural}, as summarized in Algorithm~1 in the appendix. More specifically, we learn the parameters for each module, $\Theta_F$, $\Theta_\pi$ and $\Theta_h$. The parameter sets are learned sequentially in this order as each learned parameter set is needed in the subsequent module. For instance, the barrier labels are generated by simulating backup maneuvers through the learned dynamics, which must therefore exist first. Throughout, we will denote $\theta:=(\Theta_F,\Theta_\pi,\Theta_h)$ as the collective parameter set. $F_\theta$, $\pi_\theta$ and $h_\theta$ always denote the corresponding module after training, frozen. Objects assembled from those modules inherit the subscript, so the integral law built from $\pi_\theta$ and $F_\theta$ is written $\phi_\theta$ and the extended barrier is written $h_{e,\theta}$. 
\begin{remark}[on the sequential nature of the learning approach]
    Each component is verifiable on its own against a measurable criterion, the error each contributes propagates analytically (Theorem~\ref{thm:app_error}), and the training objectives do not compete, because safety is a constraint of the online program rather than a term inside a differentiable loss.
\end{remark}

\subsection{Nonlinear system identification\label{subsec:nonlinear_system_identification}}

System identification refers to estimating or deriving, analytically or from data, a functional form for the given system such that it can be differentiated or placed inside an optimizer. 

We use Neural ODEs \citep{chen2018neural_ode} to learn $F$ in \eqref{eqn:nonlinear_system} from a dataset $\mathcal{D}_F = \{ \Gamma_1, \dots, \Gamma_{N_\tau}\}$, where each trajectory $\Gamma_j$ is a sequence $\{(\bx_0, \bu_0), \dots,(\bx_N, \bu_N)\}$ with $\bx_k = \bx(t_k)$, $\bu_k = \bu(t_k)$. The dataset is produced by open loop excitation rather than by logging a controller: initial states are drawn uniformly, actuator values are drawn uniformly from an admissible set and held piecewise constant with small jitter, and the resulting short trajectories are recorded at the control period $\Delta t$. Exciting the input box directly is what makes $\partial F/\partial\bu$ identifiable.

Our simulation studies read the labels $\dot\bx$ from the plant directly, whereas on hardware they would be differenced from the recorded states or the first term of \eqref{eqn:loss_nonlinear_system} dropped in exchange for the second, which needs only states sampled at $\Delta t$. The recorded data form two sets, isolated transitions $(\bx,\bu,\dot\bx)$ and short trajectories $\Gamma_j$, and we fit $\Theta_F$ by minimizing
\begin{align}
    \mathcal{L}_F(\Theta_F) = \sum_{(\bx,\bu)}\big\|F_{\Theta_F}(\bx,\bu)-\dot\bx\big\|^2_{W_1}
    \;+\;\lambda\sum_{j=1}^{N_\tau}\sum_{k=1}^{K}\big\|\hat\bx^j_k-\bx^j_k\big\|^2_{W_2},
    \label{eqn:loss_nonlinear_system}
\end{align}
where $\hat\bx^j_k=\Phi_{\Theta_F}\big(\hat\bx^j_{k-1},\bu^j_{k-1},\Delta t\big)$ with $\hat\bx^j_0=\bx^j_0$, diagonal normalizations $W_1,W_2$ and weight $\lambda>0$, where $\Phi_{\Theta_F}$ is the one step transition map obtained by integrating $F_{\Theta_F}$ over $\Delta t$ with a fourth order Runge-Kutta scheme. The second term feeds the model its own predictions, so it penalizes drift accumulated over $K$ steps rather than one step error. The transition map is used twice in the paper, here and in Section~\ref{subsec:neural_icbfs}, where it is the simulator through which backup maneuvers are rolled out to generate the barrier labels.

\subsection{Imitation learning for the nominal controller\label{subsec:imitation_nonlinear_mpc}}

Several expert nominal controllers exist but most unfortunately are intractable to run in realtime. It is much easier to learn a lighter weight controller that serves a realistic copy of the expert. 

We start by generating
a dataset $\mathcal{D}_c:=\{(\bx^\star_i, \bu_i^\star)\}_{i=1}^{N_c}$ by rolling the expert out on an obstacle free instance of the task from randomized initial states, with a fixed fraction of the executed actions perturbed so that the recorded states leave the expert's own closed loop distribution. That perturbation matters here more than in ordinary behavior cloning, because the filter will itself drive the system off that distribution whenever it intervenes. 

We then fit $\Theta_\pi$ by minimizing $\mathcal{L}_c(\Theta_\pi)=\mathbb{E}_{(\bx^\star,\bu^\star)\sim \mathcal{D}_c}\big[\ell\big(\pi_{\Theta_\pi}(\bx^\star),\bu^\star\big)\big]$, where $\ell$ is a per sample discrepancy between the predicted and the demonstrated action, in our case a squared error normalized by the actuator ranges.
Cloning an unconstrained expert also avoids fitting the sharp corners of a saturated policy, which feedforward networks approximate poorly, and the filter would be required in unseen states in any case. The nominal controller is moreover consumed through its Jacobian rather than its output, since $\partial\pi_\theta/\partial\bx$ is what builds the integral law. Following Remark 3 of \citet{ames2020integral_cbf} with $\pi_\theta$ in the role of $k$, that law is
\begin{align}
\phi_\theta(\bx,\bu)
=
{\frac{\partial \pi_{\theta}}{\partial \bx}(\bx)\,F_{\theta}(\bx,\bu) + \frac{\alpha_\phi}{2} \left( \pi_{\theta}(\bx)-\bu \right)},
\label{eqn:phi_definition}
\end{align}
with gain $\alpha_\phi>0$, the Jacobian being evaluated by automatic differentiation through $\pi_\theta$. Note that $\phi_\theta$ depends on two of the three blocks, $\Theta_\pi$ through $\pi_\theta$ and $\Theta_F$ through $F_\theta$. An external planner may be substituted for $\pi_\theta$, in which case $\phi_\theta=\tfrac{\alpha_\phi}{2}(\bu_{\text{nom}}-\bu)$, and Section~\ref{subsec:racing} reports both.

\begin{lemma}[Tracking error dynamics induced by $\phi$]\label{lem:tracking}
Let $F_\theta$ and $\pi_\theta$ be continuously differentiable and consider the nominal flow $\dot{\bx}=F_\theta(\bx,\bu)$ under the unfiltered integral law $\dot{\bu}=\phi_\theta(\bx,\bu)$ of \eqref{eqn:phi_definition}. Then $\be := \bu-\pi_\theta(\bx)$ satisfies $\dot{\be}=-\tfrac{\alpha_\phi}{2}\be$, so $\|\be(t)\|=\|\be(0)\|e^{-\alpha_\phi t/2}$. Under the filtered law $\dot{\bu}=\phi_\theta+\bv^\star$ the same cancellation leaves $\dot{\be}=-\tfrac{\alpha_\phi}{2}\be+\bv^\star$.
\end{lemma}
\begin{proof}
$\dot{\be}=\dot{\bu}-\frac{\partial\pi_\theta}{\partial\bx}\dot{\bx}=\frac{\partial\pi_\theta}{\partial\bx}F_\theta(\bx,\bu)+\frac{\alpha_\phi}{2}(\pi_\theta(\bx)-\bu)-\frac{\partial\pi_\theta}{\partial\bx}F_\theta(\bx,\bu)=-\frac{\alpha_\phi}{2}\be$, since the Jacobian terms cancel. The cancellation is exact under $F_\theta$, which is the model \eqref{eqn:phi_definition} is built from, and holds for the true plant only up to the mismatch $\|F-F_\theta\|$.
\end{proof}

\subsection{Viability labeled Neural ICBFs\label{subsec:neural_icbfs}}
We learn a state barrier $h_\theta(\bx):\mathcal X\rightarrow\mathbb R$ and construct the extended barrier in its general, non affine form,
\begin{align}
h_{e,\theta}(\bx,\bu) \;=\; \frac{\partial h_\theta}{\partial \bx}(\bx)\,F_\theta(\bx,\bu) \;+\; \alpha\, h_\theta(\bx),
\qquad
p_\theta(\bx,\bu)^{\top} = \frac{\partial h_{e,\theta}}{\partial \bu} = \frac{\partial h_\theta}{\partial\bx}\frac{\partial F_\theta}{\partial \bu},
\label{eqn:he_general}
\end{align}
which reduces to $p_\theta^{\top}=\frac{\partial h_\theta}{\partial\bx}g(\bx)$ when the dynamics happen to be control affine. Both derivatives in \eqref{eqn:he_general} are taken by automatic differentiation through the learned models, so no closed form for $F$ or $h$ is required anywhere. The network is trained by regression on labeled states,
\begin{align}
\mathcal L_h(\Theta_h) = \sum_i w_i\left( h_{\Theta_h}(\bx_i)-y_i \right)^2,
\label{eqn:loss_h}
\end{align}
and everything turns on how $y_i$ is defined. A geometric label, $y_i=+1$ outside the obstacles and $-1$ inside, describes where the system is at an instant rather than whether the constraint can still be met, so we replace it with a viability label defined as follows. Fix a finite library $\mathcal{M}$ of backup maneuvers and a horizon $T$. Simulate each $\mu\in\mathcal{M}$ from $\bx_i$ through the learned dynamics $F_\theta$ with the actuator model in the loop, so that steering slew and motor lag are respected, and let $c_\mu(\bx_i)$ be the minimum clearance to the constraint along that rollout. The label is
\begin{align}
y_i \;=\; \tanh\!\Big(\tfrac{1}{\kappa}\max_{\mu\in\mathcal{M}} \,c_\mu(\bx_i)\Big),
\label{eqn:viability_label}
\end{align}
with scale $\kappa>0$. A state is safe exactly when some admissible maneuver keeps the system clear for $T$ seconds, which is a finite horizon, finite library inner approximation of the viability kernel. Two properties follow immediately and distinguish \eqref{eqn:viability_label} from a worst case test over a single fixed policy. Enlarging $\mathcal{M}$ can only increase the maximum, so the labeled safe set grows monotonically toward the true viability kernel and the statistic converges as $|\mathcal{M}|$ grows, and a state inside the constraint has $c_\mu<0$ for every $\mu$, so that no geometrically unsafe state is ever labeled safe. Both properties are verified empirically in Section~\ref{subsec:viability}, where they also serve to fix the horizon and the library size.

\textbf{Enforcement margin.} Because $h_\theta$ is fitted from finite data, we enforce $h_\theta-\eta\ge0$ for a margin $\eta>0$ rather than $h_\theta\ge0$, and choose $\eta$ as the smallest value at which the measured false safe rate on held out labeled states is zero. Additive disturbances can be handled with the DO-ICBF of \citet{zinage2023disturbance}.

\subsection{The Resultant Program\label{sec:implementation}}
Let $h_{\theta,1},\dots,h_{\theta,M}$ be the active state barriers, one per nearby constraint, and let $p_i,q_i$ be the corresponding ICBF row and right hand side built from \eqref{eqn:he_general}. The safe input follows the integral law $\dot{\bu} = \phi_\theta(\bx,\bu) + \bv^\star(\bx,\bu)$ with
\begin{align}
(\bv^\star,\bs^\star) \;=\; \underset{\bv\in\mathbb{R}^m,\;\bs\in\mathbb{R}^M_{\ge0}}{\operatorname{argmin}} \; \|W\bv\|^2 + \rho\|\bs\|^2
\quad \text{s.t.}\quad
P\bv + \bs \;\ge\; Q,
\qquad
\bv \in \mathcal{V}(\bx,\bu),
\label{eqn:v_star}
\end{align}
where $P$ stacks the rows $p_i^{\top}$, $Q$ the values $q_i$, $W\succ0$ weights the input channels by their physical rates, $\rho>0$ penalizes slack, and $\mathcal{V}(\bx,\bu)$ is the space of auxiliary inputs that keep every channel within both its rate limit and its actuator range, written out in \eqref{eqn:vbox}.

In $P\bv + \bs \;\ge\; Q$, each row, $p_i^{\top}=\frac{\partial h_{\theta,i}}{\partial\bx}\frac{\partial F_\theta}{\partial\bu}$ is a barrier gradient contracted with the input Jacobian of the dynamics, involving $h_\theta$ and $F_\theta$. Substituting $h_{e,\theta,i}$ for $h$ in Definition~\ref{def:icbf} gives,
\begin{align}
q_i = -\Big(\frac{\partial h_{e,\theta,i}}{\partial\bx}F_\theta \;+\; p_i^{\top}\phi_\theta \;+\; \gamma\, h_{e,\theta,i}\Big),
\label{eqn:pq_rows}
\end{align}
with $\gamma>0$ a linear class $\mathcal{K}$ gain on the extended barrier, distinct from the $\alpha$ of \eqref{eqn:he_general} and from the actuator gain $\gamma_u$. It adds the drift of the extended barrier along the learned dynamics to the term $p_i^{\top}\phi_\theta$, which is the only place where the nominal controller acts, since $\phi_\theta$ is assembled from $\partial\pi_\theta/\partial\bx$ by \eqref{eqn:phi_definition}. The barrier and the nominal controller are therefore read through their derivatives alone, while the dynamics enter both directly, in the drift, and through the input Jacobian that makes the row nonzero. 

What \eqref{eqn:v_star} returns is a rate rather than an input, because the ICBF construction lifted $\bu$ into the state of the augmented system \eqref{eqn:integrated_augmeted_system}. The command the actuators receive is obtained by integrating that rate across one control period,
\begin{align}
\bu(t+\Delta t) = \bu(t) + \big(\phi_\theta(\bx,\bu)+\bv^\star\big)\Delta t,
\quad \bu(0) = \pi_\theta(\bx^0),
\label{eqn:u_safe_neural}
\end{align}
and $\pi_\theta$ supplies an admissible starting input because its output layer is bounded and scaled to the actuator ranges.

\section{Feasibility, Invariance, and Error Analysis}

Forward invariance is enabled assuming an admissible start provided by $\bu(0)=\pi_\theta(\bx^0)$. The construction also recovers from a start that is outside $\mathcal{U}$ but not far outside it, as part (ii) of Proposition~\ref{prop:feasible} makes precise.

\begin{prop}[Feasibility and input invariance]\label{prop:feasible}
Let $\gamma_u>0$ be the actuator gain and, for each channel $j$, let $R_j>0$ be its rate limit and $u_j^{\max}>0$ its range, so that the input constraints of \eqref{eqn:v_star} are carried by
\begin{align}
\mathcal{V}(\bx,\bu)=\big\{\bv:\; \underline{v}_j \le v_j \le \bar v_j \;\;\forall j\big\},\quad
\begin{aligned}
\underline{v}_j &= \max\{-R_j,\,-\gamma_u(u_j+u_j^{\max})\}-\phi_{\theta,j},\\[-1pt]
\bar v_j &= \min\{R_j,\,\gamma_u(u_j^{\max}-u_j)\}-\phi_{\theta,j},
\end{aligned}
\label{eqn:vbox}
\end{align}
whose second arguments are the ICBF conditions for the actuator range barriers $h_{u,j}^{\pm}=u_j^{\max}\mp u_j$. Write $\mathcal{U}_+=\{\bu:|u_j|\le u_j^{\max}+R_j/\gamma_u\ \text{for all }j\}\supseteq\mathcal{U}$, and suppose the closed loop $\dot{\bu}=\phi_\theta+\bv^\star(\bx,\bu)$ admits a solution on the interval considered. Then, for every $\bx\in\mathcal{X}$, every $\bu\in\mathcal{U}_+$ and every finite set of barrier rows, the following three statements hold.
\emph{(i)} The set $\mathcal{V}(\bx,\bu)$ is nonempty, the feasible set of \eqref{eqn:v_star} is nonempty, and the minimizer is unique.
\emph{(ii)} The realized rate $\dot{\bu}=\phi_\theta+\bv^\star$ respects every actuator rate limit, $|\dot u_j|\le R_j$, and renders both $\mathcal{U}$ and $\mathcal{U}_+$ forward invariant. On $\mathcal{U}_+\setminus\mathcal{U}$ it returns every violating channel to its range, because $u_j>u_j^{\max}$ forces $\dot u_j\le\gamma_u(u_j^{\max}-u_j)<0$.
\emph{(iii)} No input constraint is relaxed, because $\bs$ appears only in the barrier rows.
\end{prop}

The full argument is given in Appendix A.1. Proposition~\ref{prop:feasible} is what separates this filter from a soft penalty formulation: state safety is approximate, because $h_\theta$ is learned, but input feasibility is exact and is never purchased at the expense of a barrier row. Section~\ref{subsec:discussion} shows the empirical counterpart, and Section~\ref{subsec:racing} reports that the program was feasible at all $74{,}590$ control steps of the racing evaluation.

We now bound the gap between the input produced from $(F_\theta,\pi_\theta,h_\theta)$ and the input the same construction would produce from $(F,k,h)$. Write $\varepsilon_F,\varepsilon_{F'}$ for the sup norm errors of $F_\theta$ and of $\partial F_\theta/\partial\bu$, $\varepsilon_k,\varepsilon_{k'}$ for those of $\pi_\theta$ and $\partial\pi_\theta/\partial\bx$, and $\varepsilon_p,\varepsilon_q$ for the induced errors in the ICBF row and right hand side.

\begin{assumption}\label{as:reg}
 On a compact operating set $\Omega\subset\mathcal{X}\times\mathcal{U}$ the functions $F,k$ and $F_\theta,\pi_\theta$ are continuously differentiable and $h,h_\theta$ are twice continuously differentiable, so that $h_e$ and $h_{e,\theta}$ are continuously differentiable, with $\|F\|\le B_F$, $\|\phi\|,\|\phi_\theta\|\le B_\phi$ and $\|\partial h_{e,\theta}/\partial\bx\|\le B_{h'}$, the policy $\pi_\theta$ is $L_\pi$ Lipschitz, the class $\mathcal{K}$ function $\gamma$ of Definition~\ref{def:icbf} is $L_\gamma$ Lipschitz, and the ICBF is regular in the sense that $0<\sigma\le\|p\|,\|p_\theta\|\le\bar p$ with $[q]_+,[q_\theta]_+\le\bar q$.
\end{assumption}

\begin{theorem}\label{thm:app_error}
     Let $\bu^\star_\theta$ and $\bu^\star$ be obtained by one Euler step of the integral law from the same $(\bx,\bu)\in\Omega$, the first built from $(F_\theta,\pi_\theta,h_\theta)$ and the second from $(F,k,h)$, with $\bv^\star$ in each case the solution of the one row ICBF program \eqref{eqn:qp_icbf}. Under Assumption~\ref{as:reg},
    \begin{align}
        \|\bu^\star_\theta-\bu^\star\|\;\leq\;\Delta t\Big[\underbrace{B_F\,\varepsilon_{k'}+L_\pi\,\varepsilon_F+\tfrac{\alpha_\phi}{2}\varepsilon_k}_{\text{integral law}}
        \;+\;\underbrace{\tfrac{\bar p}{\sigma^{2}}\,\varepsilon_q+\Big(\tfrac{\bar q}{\sigma^{2}}+\tfrac{2\bar p^{2}\bar q}{\sigma^{4}}\Big)\varepsilon_p}_{\text{QP solution map}}\Big],
        \label{eqn:thm_eqn_ineq}
    \end{align}
    and the row errors obey $\varepsilon_p\le \big\|\tfrac{\partial F}{\partial\bu}\big\|\,\varepsilon_{h'}+\big\|\nabla h_\theta\big\|\,\varepsilon_{F'}$ and $\varepsilon_q\le L_\gamma\varepsilon_{h_e}+B_{h'}\varepsilon_F+B_F\varepsilon_{h_e'}+\bar p\,\varepsilon_\phi+B_\phi\,\varepsilon_p$, where $\varepsilon_{h'}$ and $\varepsilon_{h_e'}$ are the sup norm gradient errors of $h_\theta$ and $h_{e,\theta}$, and $\varepsilon_{h_e}$ and $\varepsilon_\phi$ the sup norm errors of $h_{e,\theta}$ and $\phi_\theta$ on $\Omega$.
\end{theorem}
The full proof is given in Appendix A.2. The theorem is stated for the one row program \eqref{eqn:qp_icbf} of Theorem~\ref{thm:icbf}, which is the setting in which $\sigma$ has its meaning. The deployed program \eqref{eqn:v_star} adds the input box, the slack variables and one row per nearby constraint, and although its solution map remains Lipschitz in the problem data because the objective is strongly convex, the governing constant is then a Hoffman constant of the active system rather than $\sigma$, and it is not uniform across changes of active set. We therefore do not claim the explicit constants of \eqref{eqn:thm_eqn_ineq} for that program.

\section{Results}
\label{sec:results}
We report four studies, beginning in Section~\ref{subsec:discussion} with two low dimensional systems and the comparison with SABLAS, continuing in Section~\ref{subsec:viability} with viability labeling in isolation, and turning in Section~\ref{subsec:racing} to the agile racing evaluation that the method targets and to its behaviour when the plant departs from the friction the models were trained on. All code is implemented in PyTorch and all filter timings are on a single CPU core.

\subsection{On low dimensional systems\label{subsec:discussion}}
The same pipeline was first validated on two low dimensional systems, a control affine kinematic bicycle and a quadrotor carrying a suspended pendulum \citep{beard2008quadrotor}, each avoiding $M$ obstacles on the way to a goal under a nonlinear MPC expert. On both it stays within $1\%$ of the nonlinear MPC cost at roughly a seventh of its runtime, because the program \eqref{eqn:v_star} deviates minimally from the unconstrained solution. The full comparison, together with two multi agent navigation methods included as runtime reference points, is given in the supplement. The remainder of this section concerns the choice that separates this framework from a jointly trained filter.

\begin{wraptable}{r}{0.5\textwidth}
\small
\centering
\resizebox{0.5\textwidth}{!}{
\begin{tabular}{lcccc}
\toprule
Method & Safety & Goal & Input compliance & Cost \\
\midrule
\textbf{Ours} & \textbf{100\%} & $71\%$ & \textbf{100\%} & $16758$ \\
SABLAS & $98.49\%$ & \textbf{89\%} & $53.38\%$ & \textbf{11474} \\
\bottomrule
\end{tabular}
}
\caption{Double integrator benchmark of \citet{qin2022sablas}, 100 random test episodes. Input compliance is the fraction of episodes on which the actuator constraint held throughout.}
\label{tab:sablas_comparison}
\end{wraptable}

Table~\ref{tab:sablas_comparison} compares against SABLAS \citep{qin2022sablas} on its own double integrator environment and makes the cost of the two architectures explicit. SABLAS reaches the goal more often and at lower cost, which is what joint training of the policy and the certificate enables. It satisfies the input constraint on $53.38\%$ of episodes, because that constraint is a penalty term competing with the task objective. Our filter satisfies it on every episode, for the structural reason given in Proposition~\ref{prop:feasible}: the input constraint is a bound on the decision variable, not a term in an objective, so there is nothing for the optimizer to trade it against. However, it comes at a cost of reduction in goal rate while maintaining higher safety with input satisfaction.

\subsection{Viability labeling\label{subsec:viability}}
\begin{figure}[t]
\centering
\includegraphics[width=0.97\textwidth]{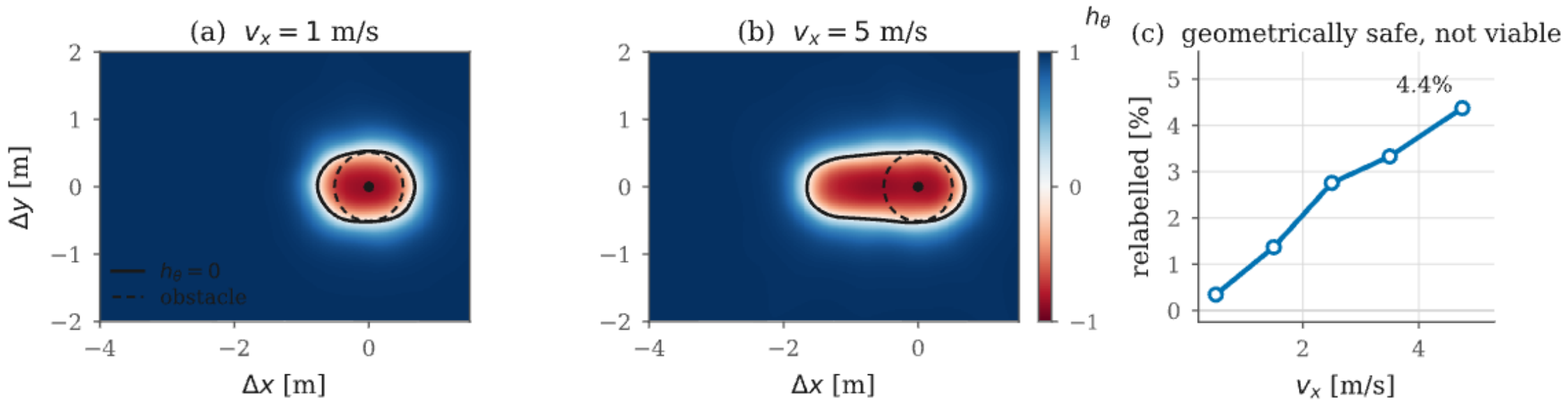}
\caption{What the viability labeled barrier represents. (left, middle) The learned obstacle barrier $h_\theta$ in the body frame at two speeds, with its zero level set and the geometric obstacle. At $1$\,m/s the two nearly coincide, whereas at $5$\,m/s the unsafe set extends roughly $1.5$\,m along the approach direction, covering states clear of the obstacle that can no longer avoid it under the actuator limits. (c) Fraction of geometrically safe states that viability relabels unsafe, by speed: the signature of a viability set rather than a smoothed obstacle indicator.}
\label{fig:viability}
\vspace{-12pt}
\end{figure}

We sample $12{,}000$ states around an obstacle, with heading, speed, slip and yaw rate drawn over the operating envelope, and compare the geometric label with the viability label \eqref{eqn:viability_label} computed through $F_\theta$ with a library of $|\mathcal{M}|=22$ backup maneuvers (three brake levels crossed with six steering policies, plus four full throttle swerves) over $T=1.5$\,s.

\textbf{Convergence of the labels.} Both limits that define the label have converged at the deployed setting. Lengthening the horizon past $0.5$\,s does not change the labels. The relabeled fraction is $2.52\%$ at $0.3$\,s and $2.60\%$ from $0.5$\,s onward, because every backup maneuver has brought the vehicle to rest by then. Enlarging the library moves the fraction monotonically downward, from $6.53\%$ with a single maneuver to $2.61\%$ at eighteen, beyond which it stops changing, so the labeled safe set is stable well before the library is exhausted. No geometrically unsafe state was ever labeled safe, in agreement with the argument in Section~\ref{subsec:neural_icbfs}. A worst case test over a single fixed policy has no such limit, since lengthening the horizon or widening the heading sample can only add unsafe states, so the fraction it reports is an artifact of where the sweep was truncated.

\textbf{Size of the correction, and its dependence on speed.} Over the whole sample, geometry marks $8.57\%$ of states unsafe and viability marks $10.95\%$, so $2.60\%$ of the geometrically safe states are already committed. Fig.~\ref{fig:viability}(c) shows that this correction is concentrated at speed: it is $0.34\%$ below $1$\,m/s, where the vehicle stops within its own footprint, and rises monotonically to $4.37\%$ above $4$\,m/s, a factor of thirteen. A barrier trained on geometric labels is therefore not merely conservative at racing speed but wrong, and it is most wrong exactly where the consequences are worst.

\textbf{Enforcement margin.} We enforce $h_\theta\ge\eta$ and pick $\eta$ by the false safe rate on held out labeled states, the fraction of unsafe states the network scores above the threshold. That rate is $1.70\%$, $0.076\%$ and $0.00\%$ at $\eta=0$, $0.1$ and $0.2$ for the obstacle barrier, and $1.81\%$, $0.038\%$ and $0.00\%$ for the wall, so we deploy $\eta=0.2$, the smallest tested value at which no held out unsafe state is scored safe.

\subsection{Agile racing with a 6-D dynamic bicycle\label{subsec:racing}}
\textbf{Setup.} We apply the same pipeline to a non control-affine 6-D dynamic single track model with F1TENTH parameters \citep{okelly2020f1tenth}, $\bx=(x,y,\psi,v_x,v_y,r)$ and $\bu=(a_x,\delta)$, with saturating lateral tire forces, a steering slew limit and first order motor lag ($\Delta t=0.02$\,s, RK4). The car races through a cluttered $50\,\mathrm{m}\times 4\,\mathrm{m}$ track at $3.5$ and $4.0$\,m/s with $\mu=1.0$. \emph{Learned components.} $F_\theta$, trained with \eqref{eqn:loss_nonlinear_system} over 10 step RK4 rollouts, has a median $1$\,s open loop position error of $8.5$\,cm, or $14.1$\,cm at the 95th percentile. More importantly for the filter, its input Jacobian $\partial F_\theta/\partial\bu$ has median cosine similarity $0.99998$ to the true Jacobian and median relative error $8.4\%$. The expert is a GPU MPPI planner \citep{williams2016mppi} on an empty track, and $\pi_\theta$ takes the target speed as an additional input. Separate barriers are learned for obstacles and for the wall, the opposite wall following by mirror symmetry, with the viability labels \eqref{eqn:viability_label}, reaching $99.7\%$ and $99.4\%$ sign accuracy. \emph{Filter.} Program \eqref{eqn:v_star} carries one row per nearby obstacle and wall, a grip row $(0.8\,\mu g)^2-a_y(\bx,\bu)^2\geq 0$ on the lateral acceleration predicted by $F_\theta$, treated as a state-input ICBF in the sense of Definition~\ref{def:icbf}, together with the input box \eqref{eqn:vbox}.

\textbf{Baselines and protocol.} We compare MPPI alone with $K=50$ samples, an analytical HOCBF-QP filter \citep{xiao2022hocbf} on MPPI, and the Neural ICBF with either $\pi_\theta$ or MPPI as nominal, on 200 layouts never used for tuning and with every parameter frozen beforehand. \emph{Unsafe} means collision or leaving the track.

\begin{wraptable}{R}{0.5\textwidth}
\centering
\footnotesize
\setlength{\tabcolsep}{5pt}
\resizebox{0.5\textwidth}{!}{
\begin{tabular}{clcccccc}
\toprule
\makecell{$v$\\(m/s)} & Method & Success & Collision & \makecell{Out of\\bounds} & \textbf{Unsafe} & Stall & Timeout \\
\midrule
\multirow{4}{*}{3.5} & MPPI                      & 172 & 0  & 1 & \textbf{1}  & 27 & 0 \\
 & HOCBF-QP                                      & 122 & 58 & 0 & \textbf{58} & 14 & 6 \\
 & Ours, $\pi_\theta$                    & 162 & 3  & 0 & \textbf{3}  & 35 & 0 \\
 & Ours, MPPI                            & 166 & 0  & 0 & \textbf{0}  & 34 & 0 \\
\midrule
\multirow{4}{*}{4.0} & MPPI                      & 168 & 7  & 5 & \textbf{12} & 20 & 0 \\
 & HOCBF-QP                                      & 55  & 69 & 1 & \textbf{70} & 18 & 57 \\
 & Ours, $\pi_\theta$                     & 155 & 1  & 0 & \textbf{1}  & 44 & 0 \\
 & Ours, MPPI                             & 169 & 0  & 0 & \textbf{0}  & 31 & 0 \\
\bottomrule
\end{tabular}
}
\caption{Agile racing on 200 held out layouts, in episode counts, with every row summing to 200. MPPI is the planner alone, HOCBF-QP is the analytical filter of \citet{xiao2022hocbf} applied to it, and Ours is the Neural ICBF with either $\pi_\theta$ or MPPI as the nominal controller. Unsafe is the sum of the collision and out of bounds columns.}
\label{tab:racing}
\vspace{-5pt}
\end{wraptable}

\textbf{Results.} Table~\ref{tab:racing} summarizes the outcomes. At $4.0$\,m/s MPPI alone is unsafe in 12 of 200 episodes, while the Neural ICBF is unsafe in 1 with $\pi_\theta$ as nominal and in 0 with MPPI as nominal, the latter at an unchanged success rate of $84.5\%$ against $84.0\%$. At $3.5$\,m/s MPPI alone is already nearly crash free, 1 of 200, and the filter matches it. The analytical HOCBF-QP is unsafe in $29$ to $35\%$ of episodes because it enforces a barrier certified for a kinematic model, and the vehicle obeys a dynamic model with tire saturation and actuator lag, so the certificate it enforces is void under the dynamics actually in force. Our learned filter avoids this because its barrier rows are differentiated through $F_\theta$, which was fitted to those dynamics.

The filter runs in $7.2$ to $7.9$\,ms per step on a single uncontended CPU core, or $8.7$ to $9.2$\,ms at the 95th percentile, so it closes the loop above $100$\,Hz without a GPU even though every row is differentiated through two networks at run time. Across the $120$ episodes and $74{,}590$ control steps on which solver diagnostics were recorded, \eqref{eqn:v_star} was feasible at every step, consistent with Proposition~\ref{prop:feasible}. 

Note that success is $3$ to $6$ points lower with $\pi_\theta$ as the nominal controller (due to the vehicle stalling), which is expected of a policy trained on an empty track and obstacle unaware by construction. The configuration we would deploy is therefore the filter on top of an obstacle aware planner (\textit{e.g.} MPPI).

\textbf{Scope.} labeled states are still required, so the framework reduces rather than removes the prior safety knowledge needed, and state safety is enforced only up to the penalized slack in \eqref{eqn:v_star}, in contrast to the input guarantee of Proposition~\ref{prop:feasible}. The method targets a sub limit regime in which MPPI requests more than $90\%$ of the available grip on only $4$ to $5\%$ of steps. At $6$\,m/s the front tire saturates, the steering gradient of the wall barrier changes sign, the margin $\sigma$ of Assumption~\ref{as:reg} collapses as the $\sigma^{-4}$ term of Theorem~\ref{thm:app_error} anticipates, and the one step program steers into the wall. Removing that failure calls for a predictive formulation rather than a larger library or a finer barrier.

\section{Conclusion\label{sec:conclusion}}
We gave a complete learning based realization of the ICBF safety filter for systems whose dynamics, nominal controller and barrier are all unknown, resting on viability labeling through the learned dynamics and on an online program in which actuator limits are hard and learned barrier rows are relaxable. That program is feasible at every state and keeps the actuator set forward invariant, and the filter built on it removes every unsafe episode from an agile racing task at an unchanged success rate and above $100$\,Hz on one CPU core. The friction limit, deadlock in front of obstacles, and transfer to hardware remain open.

\bibliography{main.bib}

\clearpage
\appendix
\section{Supplementary material}

\subsection{Proof of Proposition~\ref{prop:feasible}}

The three parts are proved in turn. Part (i) is a statement about the geometry of the box $\mathcal{V}(\bx,\bu)$ of \eqref{eqn:vbox} and about convexity of the program, part (ii) identifies the two bounds defining that box as rate limits and as barrier conditions for the actuator range, and part (iii) reads off the role of the slack variable. Throughout, $j$ indexes a single input channel, and all statements are made channel by channel because both $\mathcal{V}$ and $\mathcal{U}$ are products over $j$.

\begin{proof}[Proof of Proposition~\ref{prop:feasible}]

\textbf{Part (i), the box $\mathcal{V}$ is nonempty exactly on $\mathcal{U}_+$.}
By \eqref{eqn:vbox} the channel $j$ bounds are
\begin{align*}
\underline{v}_j &= \max\{\,\underbrace{-R_j}_{\text{rate}},\;\underbrace{-\gamma_u(u_j+u_j^{\max})}_{\text{range}}\,\}-\phi_{\theta,j},
&
\bar v_j &= \min\{\,\underbrace{R_j}_{\text{rate}},\;\underbrace{\gamma_u(u_j^{\max}-u_j)}_{\text{range}}\,\}-\phi_{\theta,j},
\end{align*}
so the interval $[\underline v_j,\bar v_j]$ is nonempty if and only if $\underline v_j\le\bar v_j$. The term $-\phi_{\theta,j}$ is common to both bounds and cancels, which leaves
\begin{align*}
\max\{-R_j,\,-\gamma_u(u_j+u_j^{\max})\}\;\le\;\min\{R_j,\,\gamma_u(u_j^{\max}-u_j)\}.
\end{align*}
For real numbers, $\max\{a_1,a_2\}\le\min\{b_1,b_2\}$ holds if and only if $a_i\le b_j$ for all four pairs, so the display above is equivalent to the following four inequalities, which we take one at a time.
\begin{align*}
\text{(a) rate against rate:}&& -R_j &\le R_j
&&\Longleftrightarrow&& 0\le 2R_j,\\
\text{(b) range against range:}&& -\gamma_u(u_j+u_j^{\max}) &\le \gamma_u(u_j^{\max}-u_j)
&&\Longleftrightarrow&& -u_j^{\max}\le u_j^{\max},\\
\text{(c) rate against range:}&& -R_j &\le \gamma_u(u_j^{\max}-u_j)
&&\Longleftrightarrow&& u_j\le u_j^{\max}+R_j/\gamma_u,\\
\text{(d) range against rate:}&& -\gamma_u(u_j+u_j^{\max}) &\le R_j
&&\Longleftrightarrow&& -u_j\le u_j^{\max}+R_j/\gamma_u.
\end{align*}
Inequalities (a) and (b) hold for every $\bu$, since $R_j>0$ and $u_j^{\max}>0$ by hypothesis. The remaining two are the only ones that constrain $\bu$, and together they say precisely that $|u_j|\le u_j^{\max}+R_j/\gamma_u$, which is membership of $\mathcal{U}_+$. Since each step above is an equivalence, the conclusion runs in both directions: $\mathcal{V}(\bx,\bu)$ is a nonempty box for every $\bu\in\mathcal{U}_+$, and it is empty as soon as $|u_j|>u_j^{\max}+R_j/\gamma_u$ for some channel. Hence $\mathcal{U}_+$ is not merely sufficient but is the largest set over which the claim can be made. The box is also bounded and closed, being a finite product of compact intervals, so it is compact.

\textbf{Part (i) continued, the program is feasible with a unique minimizer.}
Feasibility of \eqref{eqn:v_star} needs a pair $(\bv,\bs)$, not a $\bv$ alone. Take any $\bv\in\mathcal{V}(\bx,\bu)$, which part (i) has just supplied, and set
\begin{align*}
\bs := [\,Q-P\bv\,]_+,
\end{align*}
the positive part taken componentwise. Then $\bs\ge0$ by construction, and row by row either $Q_i-(P\bv)_i\le0$, in which case $(P\bv)_i+\bs_i=(P\bv)_i\ge Q_i$, or $Q_i-(P\bv)_i>0$, in which case $(P\bv)_i+\bs_i=Q_i$. Either way $P\bv+\bs\ge Q$, so $(\bv,\bs)$ is feasible and the feasible set is nonempty. It is also closed and convex, being the intersection of a box with finitely many halfspaces.

For uniqueness, the objective $\|W\bv\|^2+\rho\|\bs\|^2$ is a quadratic form in the stacked variable $(\bv,\bs)$ with Hessian
\begin{align*}
\nabla^2\!\begin{pmatrix}\bv\\\bs\end{pmatrix}
=\mathrm{blkdiag}\big(2W^{\top}W,\;2\rho I\big).
\end{align*}
Here $W\succ0$ gives $W^{\top}W\succ0$, and $\rho>0$ gives $\rho I\succ0$, so the Hessian is positive definite and the objective is strictly convex. A strictly convex function on a nonempty closed convex set attains its minimum at exactly one point, which proves the claim. Note that nothing in this argument referred to $P$ or $Q$, so it is independent of how many barrier rows there are and of whether the learned barrier is accurate.

\textbf{Part (ii), the realized rate respects the rate limits and keeps $\mathcal{U}$ and $\mathcal{U}_+$ invariant.}
The minimizer satisfies $\bv^\star\in\mathcal{V}(\bx,\bu)$, that is $\underline v_j\le v_j^\star\le\bar v_j$. Adding $\phi_{\theta,j}$ throughout and using $\dot u_j=\phi_{\theta,j}+v^\star_j$ cancels the $-\phi_{\theta,j}$ in the bounds and gives the two sided estimate
\begin{align*}
\max\{-R_j,\,-\gamma_u(u_j+u_j^{\max})\}\;\le\;\dot u_j\;\le\;\min\{R_j,\,\gamma_u(u_j^{\max}-u_j)\}.
\end{align*}
This single display carries both conclusions, obtained by keeping one entry of the $\max$ and of the $\min$ at a time.

Keeping the rate entries gives $-R_j\le\dot u_j\le R_j$, which is the rate limit $|\dot u_j|\le R_j$.

Keeping the range entries gives the barrier conditions. Write the two actuator range barriers as
\begin{align*}
h_{u,j}^{+}:=u_j^{\max}-u_j,\qquad h_{u,j}^{-}:=u_j^{\max}+u_j,
\end{align*}
so that $\mathcal{U}=\{\bu: h_{u,j}^{+}\ge0 \text{ and } h_{u,j}^{-}\ge0 \text{ for every } j\}$. Differentiating and substituting the range entries,
\begin{align*}
\dot h_{u,j}^{+}&=-\dot u_j\;\ge\;-\gamma_u(u_j^{\max}-u_j)\;=\;-\gamma_u h_{u,j}^{+},\\
\dot h_{u,j}^{-}&=\;\;\;\dot u_j\;\ge\;-\gamma_u(u_j+u_j^{\max})\;=\;-\gamma_u h_{u,j}^{-}.
\end{align*}
Both are the differential condition $\dot h\ge-\gamma_u h$ for a barrier function with the linear class $\mathcal{K}$ function $\gamma_u$. The comparison lemma applied to $\dot h\ge-\gamma_u h$ yields $h(t)\ge h(0)e^{-\gamma_u t}$, so $h(0)\ge0$ forces $h(t)\ge0$ for all $t$ in the interval considered. Applying this to $h_{u,j}^{\pm}$ for every $j$ gives forward invariance of $\mathcal{U}$.

For $\mathcal{U}_+$ the same argument runs with the enlarged barriers $H_{u,j}^{\pm}:=u_j^{\max}+R_j/\gamma_u\mp u_j$, whose zero superlevel sets intersect to $\mathcal{U}_+$. Using $H_{u,j}^{+}=h_{u,j}^{+}+R_j/\gamma_u$ and the range entry once more,
\begin{align*}
\dot H_{u,j}^{+}=-\dot u_j\;\ge\;-\gamma_u h_{u,j}^{+}\;=\;-\gamma_u\big(H_{u,j}^{+}-R_j/\gamma_u\big)\;=\;-\gamma_u H_{u,j}^{+}+R_j\;\ge\;-\gamma_u H_{u,j}^{+},
\end{align*}
since $R_j>0$, and symmetrically for $H_{u,j}^{-}$. So $\mathcal{U}_+$ is forward invariant as well.

Finally, suppose the input starts in $\mathcal{U}_+\setminus\mathcal{U}$, so that some channel has $u_j>u_j^{\max}$. The range entry of the upper bound then reads
\begin{align*}
\dot u_j\;\le\;\gamma_u(u_j^{\max}-u_j)\;<\;0,
\end{align*}
strictly negative because $u_j^{\max}-u_j<0$ and $\gamma_u>0$, so that channel decreases, and it continues to decrease until $u_j\le u_j^{\max}$. The case $u_j<-u_j^{\max}$ is symmetric, with the lower bound giving $\dot u_j\ge-\gamma_u(u_j+u_j^{\max})>0$. Every violating channel is therefore returned to its range, which is the recovery claim.

\textbf{Part (iii), no input constraint is relaxed.}
The slack $\bs$ enters \eqref{eqn:v_star} in exactly two places, the barrier rows $P\bv+\bs\ge Q$ and its own sign constraint $\bs\ge0$, together with the penalty $\rho\|\bs\|^2$ in the objective. The input constraint is the separate requirement $\bv\in\mathcal{V}(\bx,\bu)$, in which $\bs$ does not appear. Consequently no value of $\bs$, however large, can enlarge the set of admissible $\bv$, and the bounds used in part (ii) hold for the minimizer whatever the barrier rows demand. Relaxation is available to the learned part of the problem and is unavailable to the physical part.
\end{proof}

\subsection{Proof of Theorem~\ref{thm:app_error}}

The proof has four steps. Step 1 splits the deviation into a part caused by the integral law and a part caused by the quadratic program, so that the two sources of error can be bounded separately. Step 2 bounds the integral law part, Step 3 bounds the program part by exploiting the closed form of the one row program, and Step 4 traces the row errors $\varepsilon_p$ and $\varepsilon_q$ used in Step 3 back to the errors of the three networks. Every step uses the same elementary device, namely inserting a term that is a hybrid of the true and the learned object so that a difference of products becomes a sum of differences, each with one factor held fixed.

\begin{proof}[Proof of Theorem~\ref{thm:app_error}]

\textbf{Step 1, splitting the deviation.}
Both inputs are produced from the same $(\bx,\bu)$ by one Euler step of the integral law, the learned one using $(\phi_\theta,\bv^\star_\theta)$ and the true one using $(\phi,\bv^\star)$,
\begin{align*}
\bu^\star_\theta=\bu+\Delta t\big(\phi_\theta+\bv^\star_\theta\big),
\qquad
\bu^\star=\bu+\Delta t\big(\phi+\bv^\star\big).
\end{align*}
The common $\bu$ cancels on subtraction, and the triangle inequality separates the two sources,
\begin{align*}
\|\bu^\star_\theta-\bu^\star\|
=\Delta t\,\big\|(\phi_\theta-\phi)+(\bv^\star_\theta-\bv^\star)\big\|
\;\le\;\Delta t\Big(\underbrace{\|\phi_\theta-\phi\|}_{\text{Step 2}}+\underbrace{\|\bv^\star_\theta-\bv^\star\|}_{\text{Step 3}}\Big).
\end{align*}
The factor $\Delta t$ in front is the reason the statement is a one step result. It is also why both bracketed terms must be bounded by quantities that vanish with the model errors, which is what the next two steps establish.

\textbf{Step 2, the integral law term.}
By \eqref{eqn:phi_definition} the two integral laws are the same expression evaluated on different objects,
\begin{align*}
\phi=\frac{\partial k}{\partial\bx}F+\frac{\alpha_\phi}{2}(k-\bu),
\qquad
\phi_\theta=\frac{\partial \pi_\theta}{\partial\bx}F_\theta+\frac{\alpha_\phi}{2}(\pi_\theta-\bu).
\end{align*}
The difficulty is the leading term, in which both factors differ. Inserting the hybrid $\frac{\partial\pi_\theta}{\partial\bx}F$, which uses the learned Jacobian against the true dynamics, turns that product difference into two terms each having one factor fixed,
\begin{align*}
\phi_\theta-\phi
=\underbrace{\frac{\partial\pi_\theta}{\partial\bx}\big(F_\theta-F\big)}_{\text{dynamics error}}
+\underbrace{\Big(\frac{\partial\pi_\theta}{\partial\bx}-\frac{\partial k}{\partial\bx}\Big)F}_{\text{Jacobian error}}
+\underbrace{\frac{\alpha_\phi}{2}\big(\pi_\theta-k\big)}_{\text{policy error}}.
\end{align*}
Taking norms and applying Assumption~\ref{as:reg} to each factor in turn, with $\|\partial\pi_\theta/\partial\bx\|\le L_\pi$ because $\pi_\theta$ is $L_\pi$ Lipschitz and $\|F\|\le B_F$,
\begin{align*}
\|\phi_\theta-\phi\|
\;\le\;L_\pi\underbrace{\|F_\theta-F\|}_{\le\,\varepsilon_F}
+B_F\underbrace{\Big\|\tfrac{\partial\pi_\theta}{\partial\bx}-\tfrac{\partial k}{\partial\bx}\Big\|}_{\le\,\varepsilon_{k'}}
+\frac{\alpha_\phi}{2}\underbrace{\|\pi_\theta-k\|}_{\le\,\varepsilon_k}
\;\le\;B_F\varepsilon_{k'}+L_\pi\varepsilon_F+\tfrac{\alpha_\phi}{2}\varepsilon_k,
\end{align*}
which is the integral law group of \eqref{eqn:thm_eqn_ineq}. Each of its three terms is attributable to one learned object, so a reader can see which network is responsible for which part of the bound.

\textbf{Step 3, the quadratic program term.}
The one row program \eqref{eqn:qp_icbf} minimizes $\|\bv\|^2$ subject to the single constraint $p^{\top}\bv\ge q$, and this problem can be solved in closed form. If $q\le0$ the unconstrained minimizer $\bv=0$ is already feasible, and if $q>0$ the constraint is active and the minimum norm solution is the projection onto the constraint boundary along $p$. The two cases combine into
\begin{align*}
\bv^\star=\lambda\,p,\qquad \lambda=\frac{[q]_+}{\|p\|^2},
\end{align*}
and likewise $\bv_\theta^\star=\lambda_\theta p_\theta$ with $\lambda_\theta=[q_\theta]_+/\|p_\theta\|^2$. The deviation is again a difference of products, so we insert the hybrid $\lambda p_\theta$,
\begin{align*}
\bv^\star_\theta-\bv^\star=\lambda_\theta p_\theta-\lambda p=(\lambda_\theta-\lambda)\,p_\theta+\lambda\,(p_\theta-p),
\qquad\text{hence}\qquad
\|\bv^\star_\theta-\bv^\star\|\le|\lambda_\theta-\lambda|\,\|p_\theta\|+\lambda\,\|p_\theta-p\|.
\end{align*}
It remains to bound the scalar multiplier gap $|\lambda_\theta-\lambda|$. Splitting it so that the numerator and the denominator vary one at a time,
\begin{align*}
\lambda_\theta-\lambda
=\frac{[q_\theta]_+-[q]_+}{\|p_\theta\|^2}
+[q]_+\Big(\frac{1}{\|p_\theta\|^2}-\frac{1}{\|p\|^2}\Big).
\end{align*}
For the first term, $t\mapsto[t]_+$ is $1$ Lipschitz, so $\big|[q_\theta]_+-[q]_+\big|\le|q_\theta-q|\le\varepsilon_q$, and $\|p_\theta\|\ge\sigma$ by the regularity part of Assumption~\ref{as:reg}. For the second, $[q]_+\le\bar q$ and
\begin{align*}
\Big|\frac{1}{\|p_\theta\|^2}-\frac{1}{\|p\|^2}\Big|
=\frac{\big|\|p\|^2-\|p_\theta\|^2\big|}{\|p_\theta\|^2\|p\|^2}
=\frac{\big|\|p\|-\|p_\theta\|\big|\,\big(\|p\|+\|p_\theta\|\big)}{\|p_\theta\|^2\|p\|^2}
\;\le\;\frac{\varepsilon_p\cdot 2\bar p}{\sigma^4},
\end{align*}
where the numerator uses the reverse triangle inequality $\big|\|p\|-\|p_\theta\|\big|\le\|p-p_\theta\|\le\varepsilon_p$ together with $\|p\|,\|p_\theta\|\le\bar p$, and the denominator uses $\|p\|,\|p_\theta\|\ge\sigma$. Combining the two terms,
\begin{align*}
|\lambda_\theta-\lambda|\;\le\;\frac{\varepsilon_q}{\sigma^2}+\frac{2\bar p\,\bar q}{\sigma^4}\,\varepsilon_p.
\end{align*}
Substituting this into the earlier estimate, and using $\|p_\theta\|\le\bar p$ on the first term together with $\lambda=[q]_+/\|p\|^2\le\bar q/\sigma^2$ and $\|p_\theta-p\|\le\varepsilon_p$ on the second,
\begin{align*}
\|\bv^\star_\theta-\bv^\star\|
\;\le\;\bar p\Big(\frac{\varepsilon_q}{\sigma^2}+\frac{2\bar p\,\bar q}{\sigma^4}\varepsilon_p\Big)+\frac{\bar q}{\sigma^2}\,\varepsilon_p
\;=\;\frac{\bar p}{\sigma^{2}}\,\varepsilon_q+\Big(\frac{\bar q}{\sigma^{2}}+\frac{2\bar p^{2}\bar q}{\sigma^{4}}\Big)\varepsilon_p,
\end{align*}
which is the QP solution map group of \eqref{eqn:thm_eqn_ineq}. Adding this to Step 2 and multiplying by $\Delta t$ proves \eqref{eqn:thm_eqn_ineq}. The two appearances of $\sigma$ in the denominators arise from different places, one from inverting $\|p\|^2$ and one from differentiating that inverse, which is why the worst dependence is $\sigma^{-4}$ rather than $\sigma^{-2}$.

\textbf{Step 4, the row errors.}
Step 3 treated $\varepsilon_p$ and $\varepsilon_q$ as given. We now bound them by the network errors, so that the whole estimate rests on measurable quantities.

For the row, \eqref{eqn:he_general} gives $p^{\top}=\nabla h^{\top}\frac{\partial F}{\partial\bu}$, a product in which both factors are learned. Inserting the hybrid $\nabla h_\theta^{\top}\frac{\partial F}{\partial\bu}$,
\begin{align*}
p_\theta-p=\Big(\frac{\partial F_\theta}{\partial\bu}-\frac{\partial F}{\partial\bu}\Big)^{\!\top}\!\nabla h_\theta
+\Big(\frac{\partial F}{\partial\bu}\Big)^{\!\top}\big(\nabla h_\theta-\nabla h\big),
\qquad\text{so}\qquad
\varepsilon_p\le\Big\|\frac{\partial F}{\partial\bu}\Big\|\,\varepsilon_{h'}+\big\|\nabla h_\theta\big\|\,\varepsilon_{F'}.
\end{align*}
Only Jacobian errors appear here, not prediction errors, which is the reason Section~\ref{sec:results} validates $\partial F_\theta/\partial\bu$ directly rather than reporting rollout accuracy alone.

For the right hand side, Definition~\ref{def:icbf} applied to $h_{e}$ gives $q=-\gamma(h_e)-\frac{\partial h_e}{\partial\bx}F-p^{\top}\phi$, and we bound the three terms separately.
\begin{itemize}
\item The class $\mathcal{K}$ term is Lipschitz by Assumption~\ref{as:reg}, so $\big|\gamma(h_{e,\theta})-\gamma(h_e)\big|\le L_\gamma\big|h_{e,\theta}-h_e\big|\le L_\gamma\varepsilon_{h_e}$.
\item The drift term is a product in which both factors are learned. Inserting the hybrid $\frac{\partial h_{e,\theta}}{\partial\bx}F$,
\begin{align*}
\frac{\partial h_{e,\theta}}{\partial\bx}F_\theta-\frac{\partial h_{e}}{\partial\bx}F
=\frac{\partial h_{e,\theta}}{\partial\bx}\big(F_\theta-F\big)+\Big(\frac{\partial h_{e,\theta}}{\partial\bx}-\frac{\partial h_{e}}{\partial\bx}\Big)F,
\end{align*}
whose norm is at most $B_{h'}\varepsilon_F+B_F\varepsilon_{h_e'}$ by the bounds $\|\partial h_{e,\theta}/\partial\bx\|\le B_{h'}$ and $\|F\|\le B_F$.
\item The nominal term is handled the same way, inserting the hybrid $p_\theta^{\top}\phi$,
\begin{align*}
p_\theta^{\top}\phi_\theta-p^{\top}\phi=p_\theta^{\top}\big(\phi_\theta-\phi\big)+\big(p_\theta-p\big)^{\!\top}\phi,
\end{align*}
whose norm is at most $\bar p\,\varepsilon_\phi+B_\phi\,\varepsilon_p$ by $\|p_\theta\|\le\bar p$ and $\|\phi\|\le B_\phi$.
\end{itemize}
Summing the three contributions,
\begin{align*}
\varepsilon_q\;\le\;L_\gamma\varepsilon_{h_e}+B_{h'}\varepsilon_F+B_F\varepsilon_{h_e'}+\bar p\,\varepsilon_\phi+B_\phi\,\varepsilon_p,
\end{align*}
as claimed. The two barrier errors on the right are themselves controlled by the smoothness granted in Assumption~\ref{as:reg}, since $h_{e,\theta}$ is assembled from $h_\theta$ and $F_\theta$ by \eqref{eqn:he_general}, so that $\varepsilon_{h_e}$ is controlled by the $C^1$ error of $F_\theta$ together with the $C^1$ error of $h_\theta$, and $\varepsilon_{h_e'}$ by those errors at one order higher, which is where the $C^2$ requirement on $h_\theta$ is used. Every term in \eqref{eqn:thm_eqn_ineq} is therefore proportional to an error of one of the three networks, and the bound vanishes when all three are exact.
\end{proof}
\subsection{Low dimensional systems}
The vehicle is a control affine kinematic bicycle, $\dot{\bx}=f_1(\bx)\bu$, and must reach a goal while avoiding $M$ circular obstacles, while the quadrotor with a suspended pendulum uses the dynamics of \citet{beard2008quadrotor} with $M$ spherical obstacles. Expert trajectories come from nonlinear MPC solved with \texttt{do-mpc}. Table~\ref{tab:vehicle} reports the average time to reach the goal from $100$ sampled initial states together with the average closed loop quadratic cost in state and input. Nonlinear MPC attains the lowest cost at the highest runtime, since it solves non convex programs online, whereas our filter stays within $1\%$ of that optimum at a fraction of the runtime, because the program \eqref{eqn:v_star} deviates minimally from the unconstrained solution. ORCA-MAPF \citep{orcamapf} and CADRL \citep{cadrl} serve as runtime reference points only, since both were designed for reciprocal multi agent collision avoidance rather than as safety filters, so the cost column should not be read as a like for like comparison against them.

\begin{table}[h]
\small
\centering
\setlength{\tabcolsep}{4pt}
\begin{tabular}{|l|c|c||c|c|}
\hline
 & \multicolumn{2}{c||}{Vehicle} & \multicolumn{2}{c|}{Quadrotor-pendulum} \\
\hline
Method & Time (s) & Cost & Time (s) & Cost \\
\hline
NMPC & 5.562 & \textbf{4.583} & 10.242 & \textbf{15.232} \\
ORCA-MAPF & 1.373 & 6.839 & 2.328  & 21.281 \\
CADRL & 0.837 & 5.683 & 2.181  & 19.283 \\
Ours & \textbf{0.736} & 4.631 & \textbf{1.550}  & 15.672 \\
\hline
\end{tabular}
\caption{Time to goal and average closed loop cost on the two low dimensional systems. ORCA-MAPF and CADRL are runtime reference points rather than safety filters, so the cost column is not a like for like comparison against them.}
\label{tab:vehicle}
\end{table}

\subsection{Ablations\label{app:ablations}}

\textbf{Ablations.} Table~\ref{tab:ablation} isolates three choices on the held out layouts at $4.0$\,m/s on which MPPI alone is unsafe. Twelve layouts cannot separate outcome counts, so the discriminating quantities are the clearance reached and the use made of the relaxation. Every configuration kept the program feasible at every step, including those that degrade safety, which is Proposition~\ref{prop:feasible} at work: feasibility follows from the structure of \eqref{eqn:v_star} rather than from the quality of $h_\theta$ or the choice of gains.
Of the three, the grip row matters most, since it is the only state-input row and without it the filter may command lateral accelerations the tires cannot deliver. The smallest obstacle barrier value reached falls from $+0.29$ to $-0.05$, so the vehicle leaves the learned safe set, and slack becomes active on $2.77\%$ of steps against $0.16\%$ with the row present. The filter stays feasible and still intervenes, but it is enforcing a barrier through a model of the input response that the plant no longer obeys. The margin and the gain act on that same clearance from opposite directions. Removing the margin, $\eta=0$, reduces the smallest barrier value reached to $+0.20$, which is what the margin was introduced to buy and matches the open loop false safe rates of Section~\ref{subsec:viability}, while raising the gain from $\alpha=5$ to $\alpha=10$ halves it to $+0.14$, because a larger $\alpha$ lets the extended barrier tolerate a faster approach while $h_\theta$ is still positive. Neither changes the outcome counts on these layouts.
\input{ablation_table}

\subsection{Behavior under plant mismatch\label{app:mismatch}}

\textbf{Behavior under plant mismatch.} Every learned component assumes the friction its training data was generated at, $\mu=1.0$, so lowering the friction of the plant alone probes the regime that the split in \eqref{eqn:v_star} was designed for. On 16 held out layouts at $4.0$\,m/s with all three models left unchanged, the planner alone is unsafe on 3 layouts at $\mu=0.85$ and on 7 at $\mu=0.70$, against 0 and 1 for the filtered system, and it reaches the goal on 10 and 6 against 13 and 11. Both gaps widen as the plant departs from the models, because under mismatch it is the planner and not the filter that degrades, and at $\mu=0.70$ the safety difference approaches significance on 16 layouts alone ($p=0.07$). The mechanism is visible in the solver: slack activity, $0.16\%$ of steps at matched friction, rises by roughly an order of magnitude to $1.4\%$ and $1.0\%$, while \eqref{eqn:v_star} stays feasible at every step and no actuator limit is ever relaxed. The relaxation absorbs the model error exactly where Proposition~\ref{prop:feasible} says it must.

\subsection{Algorithm}
\begin{algorithm}[H]
\DontPrintSemicolon
\SetAlgoLined
\KwIn{expert $\pi_{\text{exp}}$, sampling distributions, backup library $\mathcal{M}$, horizon $T$, gains $\alpha,\alpha_\phi,\gamma,\gamma_u$, margin $\eta$}
\KwOut{$F_\theta$, $\pi_\theta$, $h_\theta$ and the online filter}
\textbf{Stage 1 (dynamics).} Excite the plant, collect $\mathcal{D}_F$, fit $F_\theta$ by \eqref{eqn:loss_nonlinear_system} and report the $\partial F_\theta/\partial\bu$ error\;
\textbf{Stage 2 (nominal).} Roll out $\pi_{\text{exp}}$ on the unconstrained task, collect $\mathcal{D}_c$, fit $\pi_\theta$\;
\textbf{Stage 3 (barrier).} \ForEach{sampled state $\bx_i$}{
  \ForEach{$\mu\in\mathcal{M}$}{simulate $\mu$ from $\bx_i$ through $F_\theta$ with the actuator model for $T$ and record $c_\mu(\bx_i)$\;}
  $y_i \leftarrow \tanh\!\big(\max_\mu c_\mu(\bx_i)/\kappa\big)$\;}
Fit $h_\theta$ by \eqref{eqn:loss_h}, then pick $\eta$ as the smallest margin with zero false safe rate on held out states\;
\textbf{Online.} \While{running}{
  assemble $p_i,q_i$ from \eqref{eqn:he_general} by automatic differentiation through $h_\theta$ and $F_\theta$\;
  build $\mathcal{V}(\bx,\bu)$ from the actuator ranges and rates by \eqref{eqn:vbox}\;
  solve \eqref{eqn:v_star} for $\bv^\star$, then set $\bu \leftarrow \bu + (\phi_\theta+\bv^\star)\Delta t$ and apply it\;}
\caption{Neural ICBF safety filter: offline training and online use}
\end{algorithm}

\subsection{Network architectures and training}
$F_\theta$ is a three hidden layer network of width $256$ with $\tanh$ activations, taking $(\cos\psi,\sin\psi,v_x,v_y,r,a_x,\delta)$ and returning $\dot\bx$, where smooth activations are required because the filter differentiates through it twice. $\pi_\theta$ is a four hidden layer network of width $256$ with a $\tanh$ output scaled to the actuator ranges. Both barriers are three hidden layer networks of width $256$ with a $\tanh$ output, taking the relative state and the obstacle radius. Training uses AdamW with cosine annealing from $10^{-3}$ to $10^{-5}$. Unsafe states are weighted five times in \eqref{eqn:loss_h}.

\end{document}

%% file: packages.tex
\usepackage{times}
\usepackage{algorithmic}
\usepackage[font=small]{caption}
\usepackage{graphicx}
\usepackage{adjustbox}
\usepackage{multirow}
\usepackage{pbox}
\usepackage{amsmath}
\usepackage{wrapfig}
\usepackage{booktabs}
\usepackage{makecell}
\newcommand{\bx}{{\boldsymbol{x}}}
\newcommand{\bu}{{\boldsymbol{u}}}

\newcommand{\bz}{{\boldsymbol{z}}}

\newcommand{\bv}{{\boldsymbol{v}}}

\newcommand{\be}{{\mathbf{e}}}
\newcommand{\bs}{{\boldsymbol{s}}}

\usepackage{adjustbox}
 \usepackage{multirow}
\usepackage{pbox}

\usepackage{tabu}

\usepackage{amsmath}
\usepackage{amssymb}
\usepackage{calc}
\usepackage{tcolorbox}
\newtheorem{problem}{Problem}
\newtheorem{prop}{Proposition}

\newtheorem{assumption}{Assumption}

%% file: macros.tex
\usepackage[skip=.3em plus 2pt minus 1pt, indent=0pt]{parskip}

%% file: ablation_table.tex
\begin{wraptable}{r}{0.5\textwidth}
\centering
\resizebox{0.5\textwidth}{!}{
\begin{tabular}{lcccccc}
\toprule
Configuration & Success & Unsafe & $\min h_\theta$ & QP feasible & Slack active & Interventions \\
\midrule
Default ($\eta=0.2$, $\alpha=5$, grip row) & 8/12 & 0/12 & $+0.29$ & $100\%$ & $0.16\%$ & $44\%$ \\
No enforcement margin ($\eta=0$) & 9/12 & 0/12 & $+0.20$ & $100\%$ & $0.13\%$ & $45\%$ \\
Barrier gain $\alpha=10$ & 9/12 & 0/12 & $+0.14$ & $100\%$ & $0.15\%$ & $45\%$ \\
No grip row & 7/12 & 0/12 & $\mathbf{-0.05}$ & $100\%$ & $2.77\%$ & $31\%$ \\
\bottomrule
\end{tabular}
}
\caption{Ablations on the 12 held out layouts at $4.0$\,m/s on which MPPI alone is unsafe, with MPPI as the nominal controller. $\min h_\theta$ is the smallest obstacle barrier value reached in closed loop, before the enforcement margin is subtracted, so a negative entry means the learned safe set was left. Interventions is the fraction of steps at which $\bv^\star\neq0$.}
\label{tab:ablation}
\end{wraptable}